\documentclass{article}

\usepackage{microtype}
\usepackage{graphicx}
\usepackage{subfigure}
\usepackage{booktabs} 

\usepackage{hyperref}

\usepackage[accepted]{icml2023}

\usepackage{amsmath}
\usepackage{amssymb}
\usepackage{mathtools}
\usepackage{amsthm}

\usepackage[capitalize,noabbrev]{cleveref}
\newcommand{\Ebb}{\mathbb{E}}

\newcommand{\Rbb}{\mathbb{R}}

\newcommand{\Contextspace}{\mathcal{X}}
\newcommand{\preferred}{\succ}
\newcommand{\maybepreferred}{\overset{?}{\preferred}}
\newcommand{\Actionspace}{\mathcal{A}}

\newcommand{\bestpolicy}{\hat{\pi}_T}

\newcommand{\ainset}{a \in \Actionspace}

\newcommand{\lcbr}{\underline{f_r^t}}
\newcommand{\ucbr}{\overline{f_r^t}}
\newcommand{\bonus}{\beta_t^{(r)}}
\newcommand{\lastbonus}{\beta_T^{(r)}}
\newcommand{\kernel}{\kappa}

\newcommand{\borda}{f_r}
\newcommand{\algnm}{Borda-AE}
\newcommand{\uniform}{Borda-Uniform}
\newcommand{\uucb}{Borda-UCB}
\newcommand{\xinset}{x \in \Contextspace}

\DeclareMathOperator{\argmax}{arg \max}

\DeclareMathOperator{\Bernoulli}{Bern}

\definecolor{nicegreen}{RGB}{91,226,91}

\theoremstyle{plain}
\newtheorem{theorem}{Theorem}[section]

\newtheorem{lemma}[theorem]{Lemma}

\theoremstyle{definition}

\newtheorem{assumption}[theorem]{Assumption}
\theoremstyle{remark}

\usepackage[textsize=tiny]{todonotes}

\icmltitlerunning{Kernelized Offline Contextual Dueling Bandits}

\begin{document}

\twocolumn[
\icmltitle{Kernelized Offline Contextual Dueling Bandits}

\begin{icmlauthorlist}
\icmlauthor{Viraj Mehta}{ri}
\icmlauthor{Ojash Neopane}{mld}
\icmlauthor{Vikramjeet Das}{mld}
\icmlauthor{Sen Lin}{}
\icmlauthor{Jeff Schneider}{ri,mld}
\icmlauthor{Willie Neiswanger}{stan}
\end{icmlauthorlist}

\icmlaffiliation{ri}{Robotics Institute, Carnegie Mellon University, USA}
\icmlaffiliation{mld}{Machine Learning Department, Carnegie Mellon University, USA}
\icmlaffiliation{stan}{Department of Computer Science, Stanford University, USA}

\icmlcorrespondingauthor{Viraj Mehta}{virajm@cs.cmu.edu}

\icmlkeywords{Machine Learning, ICML}

\vskip 0.3in
]

\printAffiliationsAndNotice{\icmlEqualContribution} 

\begin{abstract}
Preference-based feedback is important for many applications where direct evaluation of a reward function is not feasible. A notable recent example arises in reinforcement learning from human feedback on large language models. For many of these applications, the cost of acquiring the human feedback can be substantial or even prohibitive. In this work, we take advantage of the fact that often the agent can choose contexts at which to obtain human feedback in order to most efficiently
identify a good policy, and introduce the \emph{offline contextual dueling bandit} setting. We give an upper-confidence-bound style algorithm for this setting and prove a regret bound. We also give empirical confirmation that this method outperforms a similar strategy that uses uniformly sampled contexts.
\end{abstract} \section{Introduction}
In many decision making problems in information retrieval, question answering, clinical trials, advertising, and other fields, feedback about the performance of a particular choice is only available in the form of a preference between several options. 
Such feedback often takes the form of a user clicking on a link in the wild but can also be collected from labelers in an attempt to understand their underlying preferences and train a system to optimize them.
This is an especially useful method for collecting human feedback because humans are unreliable in giving scalarized feedback compared to the accuracy of their preferences \citep{ouyang2022training}.
Often, these settings come with some context that can provide information about the associated distribution over preferences. This could be the search term, some information about a trial participant, or the prompt in a question answering setting.

Recently, these techniques have seen large amounts of attention when popularized through reinforcement learning from human feedback (RLHF). RLHF is one of the major techniques for aligning large language models (LLMs) for use as a chat assistant or for other specialized applications after pretraining on a sequence modeling objective.
In these applications, human raters are provided with a prompt and several possible responses taken from the LLM.
They are asked to rank the human responses based on their preferences given the prompt.
This process requires a relatively large number of samples from human raters (tens of thousands) in order for the alignment process to succeed. 
This can incur large costs for the data collection.
For more specialized problems than a general chatbot assistant, the feedback required may be impractical or expensive relative to the desired application.

In many applications, the feedback is collected \emph{online} as the policy is learned. 
Under these circumstances, the contexts are typically assumed to be drawn from an (unknown) stationary probabity distribution and the agent's object is to quickly find a near optimal decision rule.
Currently, in the RLHF setting, the prompts presented to the model in order to sample responses and then to the raters are typically sampled uniformly from a response set intended to be representative of the test-time distribution.

In such cases, the aforementioned online setting does not allow us to fully take advantage of the problems structure -- we can control which prompts and responses are presented to the human raters for feedback and we are not interested in the performance of the actions chosen during labeling, just the performance of the policy at test time afterwards.
Instead of the standard contextual bandit problem it is then more appropriate to consider the so-called \emph{offline contextual} bandit \cite{char_ocbo} where we are additionally allowed to select the contexts for which we receive feedback.
By leveraging this control over this less restrictive data-generating process, we show how to select contexts and actions in order output policies with stronger optimality guarantees without the need to collect more data.

Here, we tackle the special case of pairwise feedback where a pair of actions are compared given a particular context.
Following \citet{xu2020zeroth}, we first reduce the problem of finding the optimal action given pairwise feedback to finding the action that optimizes the \emph{Borda function} given a particular context.
The Borda function is the probability that for a particular context, a selected action is preferred over another action selected uniformly at random.
We select contexts which maximize the uncertainty over the Borda `value function' and then select one action optimistically and the other uniformly.

In this work, we show that our method provably achieves suboptimality at most $O\left(\frac{L_1}{\sqrt{T}}\left(B + \Phi_T\sqrt{\log\frac{1}{\delta}}\right)\right)$  everywhere in the context space after $T$ iterations with probability $1 - \delta$ under the assumption that the Borda function is bounded by $B$ in RKHS norm. We also demonstrate on a distribution of synthetic problems that it performs well when implemented and outperforms a baseline with uniformly selected contexts and an optimistic policy as well as entirely uniform sampling. \section{Related Work}
\paragraph{Learning from Comparative Feedback}

There is a rich literature on reinforcement learning from comparative human feedback, including work by \citet{furnkranzPreferenceRL}, \citet{akhourRobustPreferenceRL} and \citet{deepRLFromHumanPreferences}. Many of these works grappled with the heightened need for sample efficiency given the cost of acquiring human feedback. In particular, \citet{deepRLFromHumanPreferences} made it feasible to use human feedback for deep reinforcement learning by training a reward model that is then used as the target for reinforcement learning. In their Atari test case, where naive deep RL would have required thousands of hours of gameplay, they were able to achieve superior performance with only 5,500 or several hours of human queries. 

More recently, methods of using comparative human feedback have gained prominence as a means of improving the performance of language models. These methods have been shown to be effective at improving stylistic continuation \citep{ziegler2019fine}, text summarization \citep{stiennon2020}, translation \citep{Kreutzer2018}, semantic parsing \citep{LawrenceAndReizler2018}, review generation \citep{Cho2018}, and evidence extraction \citep{Perez2019}. However, while effective, incorporating human feedback brought substantial costs.
For example, \citet{stiennon2020} achieved significant improvements to baseline, but needed summaries on 123,169 posts from the TL;DR dataset generated by a small team of labelers (more than 21 persons) from Upwork, Scale, and Lionbridge to train.

This heavy-resource requirement is again reflected even in later, state-of-the-art work. \citet{ouyang2022training} focused on using RLHF to improve alignment of the GPT-3 model (at 175B parameters) with human values on a variety of directions, including toxicity, hallucinations, moral opinion, and overall quality. The results are spectacular, with the 1.3B parameter InstructGPT matching the 175B GPT-3 in performance on a variety of tasks. Because the focus was ensuring representation both in the inputs to models used in real life and in the human feedback received, the team used 40 labelers and worked with a dataset of more than 100,000 examples. 

A related paper from \citet{zhu2023principled} also explores a simplified version of this problem and showed that under the strong assumption of a linear model given a known feature mapping, the policy obtained by optimizing the pessimistic MLE given a fixed dataset is provably optimal for learning in the $k$-wise comparison context. Given these strong assumptions, the authors point out that a $G$-optimal experimental design for online data collection as in \citet{soare2014best} would be maximally informative. However, these assumptions are unrealistic and do not represent the methods used in practice as reward model training is usually conducted over all layers of a deep model.
\paragraph{Dueling Bandits}

At the same time, the bandit literature has also explored the effectiveness of comparative feedback (``dueling bandit'') while considering the cost of acquiring such information. This was first studied by \citet{yue2012} in settings where comparative information is relatively easy to extract but absolute rewards (\textit{i.e.}, direct queries) are ill-defined and have no absolute scale. Later, \citet{bengs2021preferencebased} surveyed methods used in the online learning setting, where the trade off with cost of information is most acute, including those used in the online contextual dueling bandit setting by \citet{dudk2015contextual}. 
These constraints motivate a kernelized approach that can incorporate the nonlinearities in the models used in practice.

\paragraph{Offline Contextual Bandit Optimization}

When there are distinct phases of learning and then deployment, an agent can make maximal use of every example during learning to acquire information that can be exploited once deployed.

\citet{char_ocbo} introduce this idea for black-box function approximation by considering a setting where at test time the goal is to perform well on average across a context distribution while during learning the goal is to choose contexts and actions that are most useful for that goal.  
Given a reward function for each task, the authors proposed a multi-task version of Thompson sampling during the offline training phase, which allows provable regret bounds in that problem setting. We extend this setting from cardinal to ordinal rewards as is appropriate for comparative feedback.

In \citet{li2023nearoptimal}, the agent queries the states (or contexts, in a bandit setting) where the value function is most uncertain and acts optimistically. Combined with least-squares value iteration, this method leads to provable polynomial-sample convergence in the worst-case error of the value function estimate in reinforcement learning in general, and as a corollary the setting from \citet{char_ocbo} as a special case. This sets the foundation that we will adapt to the comparative feedback setting.

 \newcommand{\preferenceMatrix}{f}
\newcommand{\subopt}{\operatorname{SubOpt}}
\newcommand{\RKHS}{\mathcal{H}}
\section{Problem Setting}
\label{s:problem_setting}
In this paper, we consider a dueling variant of the so-called offline contextual bandit problem introduced in \citet{char_ocbo}.
An instance of this problem is defined by a tuple $(\Contextspace, \Actionspace, \preferenceMatrix)$ where $\Contextspace$ denotes the context space, $\Actionspace$ denotes the action space and $\preferenceMatrix: \mathcal X \times \mathcal A \times \mathcal A \rightarrow [0, 1]$ is a preference function so that $\preferenceMatrix(x, a, a')$ denotes the probability that the action $a$ is preferred to the action $a'$ when the underlying context is $x$.
We will design algorithms that operate under the following interaction protocol, which occurs for $T$ time steps.
During each time step $t \in [T]$, the agent selects a context $x_t \in \Contextspace$ and a pair of actions $a_t, a_t' \in \Actionspace$ and observes a binary random variable $R_t \sim \Bernoulli(\preferenceMatrix(x_t, a_t, a_t'))$ which equals one if $a_t$ is preferred to $a_t'$ (denoted $a_t \preferred a_t'$) and zero otherwise.

We assume that the preference function takes the following form
\begin{equation}
    \preferenceMatrix(x, a, a') = \sigma \left( r(x, a) - r(x, a') \right),
\end{equation}
where $\sigma: \Rbb \rightarrow [0, 1]$ is the \emph{link function} and $r: \mathcal X \times \mathcal A \rightarrow \Rbb$ is the reward function.
Common link functions include the logistic function, which leads to the Bradley-Terry-Luce (BTL) model \cite{bradley1952rank} as well as the Gaussian CDF \cite{thurstone1927method}.
We also place some additional assumptions on the reward function which we discuss at the end of this section. 

Our objective within this protocol is to design algorithms that are able to quickly identify policies with small suboptimality. We define the suboptimality of a policy $\pi$ as
\begin{equation}
    \subopt(\pi) = \sup_{x \in \Contextspace} \left( \sup_{a \in \Actionspace} r(x, a) - r(x, \pi(x)) \right).
\end{equation}
We remark that this notion of suboptimality is much stronger than usual notions that look at the expected suboptimality of the final policy when the contexts are sampled from some known distribution.
In contrast, the form of suboptimality we consider here looks at the worst-case context for each policy.

Before discussing the assumptions we place on the reward function, we first introduce a closely related function, called the \emph{contextual borda function} $f_r$, which generalizes the borda function introduced in by \citet{xu2020zeroth}.
The Borda function as introduced in \citet{xu2020zeroth} for dueling-choice optimization is defined as the probability that a particular action $a$ will be preferred over a random action $a'$ uniformly sampled from the action space.
We generalize this definition to the contextual setting as follows, given as $\borda: \Contextspace \times \Actionspace \to [0, 1]$ where $\borda(x, a) = \Ebb_{a'\sim U(\Actionspace)}\left[P(a \preferred a' \mid x)\right]$, where $U(\Actionspace)$ is the uniform measure over the action space. 
It is clear from the definition that $f_r$ and $r$ will have the same maximizers.

We conclude this section by discussing the structural assumptions we place on the reward function as well as the contextual Borda function.
Our first assumption restricts the reward and contextual Borda functions to be `smooth' in an underlying Reproducing Kernel Hilbert Space (RKHS).
\begin{assumption}
    Let $\kernel$ denote a Positive Semi-Definite kernel and let $\RKHS_\kernel$ denote its associated RKHS. We assume that $ \left \lVert r \right \rVert_{\kernel}, \left \lVert f_r \right \rVert_{\kernel} \leq B$, where $B$ is a known constant.
\end{assumption}
Note that this assumption is different than the standard assumption which only requires that $r$ has bounded RKHS norm.
This is due to the generality of our setting which allows for multiple different link functions. 
While this assumption is not ideal, it is difficult to bound the norm of $f_r$ given a bound on the norm of $r$.
We investigate this issue more in Section~\ref{s:rkhs_borda} where we empirically find that the norm of the Borda function is almost always smaller than the norm of the reward function.

Our second assumption relates the reward function to the contextual Borda function.
\begin{assumption}
    \label{ass:borda}
    Let $\borda^*(x) = \max_a \borda(x, a)$ and $r^*(x) = \max_a r(x, a)$. There exists a constant $L_1$ such that for every $x \in \Contextspace$, $a \in \Actionspace$ we have $\frac{1}{L_1}(r^*(x) - r(x, a)) \leq \borda^*(x) - \borda(x, a)$.
\end{assumption}
This assumption implies that differences in $r$ will cause a similar magnitude of difference in $f_r$ In fact, when $\sigma$ is Lipschitz continuous, it is sufficient for the Lipschitz constant of $\sigma$ to be at least $1/L_1$ for this condition to hold.

 \section{Method and Analysis}
\label{sec:methodandanalysis}
At a high level, our approach reduces the dueling feedback problem to contextual optimization over a single action via the \emph{contextual Borda function} introduced in Section~\ref{s:problem_setting}. 
Once reduced appropriately, we apply techniques adapted from recent work on active exploration in reinforcement learning to construct a sampling rule and policy selection rule which allows us to output a policy with provably small suboptimality.
Broadly, our sampling rule samples contexts at which there is maximum uncertainty over the Borda `value function' and then compares the optimistic action with an action sampled uniformly from the action set. 

\subsection{Estimating The Contextual Borda Function}
By design, we can easily estimate the contextual Borda function from data of the form $\{x_t, a_t \maybepreferred a'_t\}$, where the contexts $x_t$ and first actions $a_t$ are arbitrary and the second actions $a_t'$ are uniformly selected. 
In this work, we model the contextual Borda function using standard kernelized ridge regression (KRR) \citep{rasmussen2006gaussian}.
The key feature of KRR is that besides an estimate of the contextual Borda function after $t$ observations
$\mu_t(x, a)$, we can also estimate the uncertainty over the prediction $\sigma_t(x, a)$ using standard results; under the assumptions in Section~\ref{s:problem_setting} and given an value for $\beta$ appropriate chosen for our confidence level, we can bound $|f_r(x, a) - \mu_t(x, a)| \leq \beta \sigma_t(x, a)$ with the desired confidence.

\subsection{Selecting Contexts and Actions}
Our sampling rule builds on top of the one established in \citet{li2023nearoptimal} ---put simply the rule is to sample the state with the maximum uncertainty over the value function and then act optimistically.
We will now  present our algorithm which highlights how to extend these idea to the dueling setting via the contextual Borda function $f_r$.

For now, we assume that there is a known bonus term $\bonus$ for all $t$. We can then define upper and lower confidence bounds $\ucbr(x, a) = \mu_t(x, a) + \bonus \sigma_t(x, a)$ and $\lcbr(x, a) = \mu_t(x, a) - \bonus \sigma_t(x, a)$.
Our rule is to sample a context
\begin{equation}
\label{eq:context_selection}
\begin{aligned}
    x_t \in \argmax_{\xinset}\left(\max_{\ainset} \ucbr(x, a) - \max_{\ainset}\lcbr(x, a)\right).
\end{aligned}
\end{equation}
Here, we are choosing a context that maximizes the difference between the optimistic `value function' and the pessimistic `value function' (both of which require a maximization over actions to compute). 

We then optimistically choose an action by
\begin{equation}
\label{eq:action_selection}
    a_t \in \argmax_{\ainset}\ucbr(x_t, a).
\end{equation}
After repeating this process $T$ times, we output a pessimistic policy against the tightest lower bound we can find, which is the maximizer of all our lower bounds through the optimization process. Put formally we return $\bestpolicy: \Contextspace \to \Actionspace$ such that
\begin{equation}
\label{eq:bestpolicy}
    \bestpolicy(x) \in \argmax_{a \in \Actionspace}\max_{t \leq T}\lcbr(x, a).
\end{equation}
From these pieces we construct the full algorithm, Borda-AE, which we present in Algorithm~\ref{alg:Borda-AE}. 
\begin{algorithm}[]
        \caption{\algnm}
        \label{alg:Borda-AE}
        \begin{algorithmic}[1] \STATE {\bfseries Input:} kernel function $\kernel(\cdot, \cdot)$, exploration parameters $\bonus$, number of inital data $n_0$
            \STATE Let $D_{n_0} = \{s_i, a_i \maybepreferred a'_i\}_{i=1}^{n_0}$ for $s_i, a_i, a'_i$ uniform.
            \FOR {$t=n_0 + 1,\dots,T$}
                \STATE Compute $\mu_t(\cdot, \cdot)$, $\sigma_t(\cdot, \cdot)$ using KRR.
                \STATE Choose $x_t$ according to \eqref{eq:context_selection}.
                \STATE Choose $a_t$ according to \eqref{eq:action_selection}, $a'_t\sim U(\Actionspace)$.
                \STATE Let $D_t = D_{t-1} \cup \{(x_t, a_t \maybepreferred a'_t)\}$.
            \ENDFOR
            \STATE Output a final policy $\bestpolicy$ according to \eqref{eq:bestpolicy}.
        \end{algorithmic}   
    \end{algorithm}
\subsection{Bounding the regret of \algnm}
Before proceeding with our algorithm's formal guarantees, we first introduce the \emph{maximal-information gain} which plays an important role in our results.
The maximum information gain over $t$ rounds, denoted $\Phi_t$, is defined as
\begin{equation}
    \Phi_t = \max_{A \subset \Contextspace\times\Actionspace: \left \lvert A \right\rvert = t} I(r_A + \epsilon_A; r_A),
\end{equation}
where $r_{A} = \left[ r(x) \right]_{x \in A}$ , $\epsilon_A \sim N(0, \eta^2 I)$ and $I(X; Y) = H(X) - H(X | Y)$ is the mutual information.
With this definition, we are now ready to state our result.
\begin{theorem}
    \label{thm:regret}
    Suppose we run Algorithm~\ref{alg:Borda-AE} with
    \begin{equation}
        \beta^{(r)}_t = 2B + \sqrt{2 \Phi_t + 1 + \log \left( \frac 2 \delta \right)},
    \end{equation}
    then, with probability at least $1 - \delta$, we have that
    \begin{equation}
        \subopt(\hat \pi_T) \leq O \left(  \frac{L_1}{\sqrt{T}} \left(B + \Phi_T\sqrt{\log \frac{1}{\delta}} \right)\right).
    \end{equation}
\end{theorem}
\paragraph{Proof Sketch.} 
At a high-level the proof of this result is as follows. 
First, we use standard results on KRR to show that our choice of $\beta^{(r)}$ guarantees that our confidence bands contain $f^r(x, a)$ with high probability simultaneously for all $t$ and $x, a \in \Contextspace \times \Actionspace$.
Next, we use assumption~\ref{ass:borda} to show that the suboptimality of the  pessimistic policy induced by our estimated contextual borda function is small whenever we are able to estimate the contextual borda function well.
Finally, we conclude the proof by showing that our sampling rule indeed allows us to estimate thet contextual borda function well.
\paragraph{Concrete Performance Bounds.} 
To more concretely understand the performance of our algorithm, we instantiate our results for three commonly studied kernels: the linear, squared-exponential.
For both of these settings, the scaling of the information gain is well known (see for example \cite{srinivas2009gaussian}).
In the linear setting, we have that $\Phi_T = d \log T$ leading to a bound of $O \left( \frac{L_1}{\sqrt{T}} \left( d B \log T \sqrt{ \log \frac 1 \delta} \right) \right)$.
For squared exponential kernels we have $\Phi_T = O \left( \log(T)^{d + 1} \right)$ leading to a suboptimality bound of $O \left( \frac{L_1}{\sqrt{T}} \left( B \log(T)^{d + 1} \sqrt{ \log \frac 1 \delta} \right) \right)$.

When compared to existing results for dueling bandits \cite{xu2020zeroth} as well as standard bandits \cite{chowdhury2017kernelized}, we see that our suboptimality bounds match, thus showing that our algorithm is able to achieve the same performance under a stronger performance metric.

 \section{Experiments}
\label{s:experiments}
\begin{figure*}
    \centering
    \hspace{5mm} \textbf{Time = 50} \hspace{34mm} \textbf{Time = 150} \hspace{34mm} \textbf{Time = 600} \\
    \vspace{3mm}
    \includegraphics[width=0.3\textwidth]{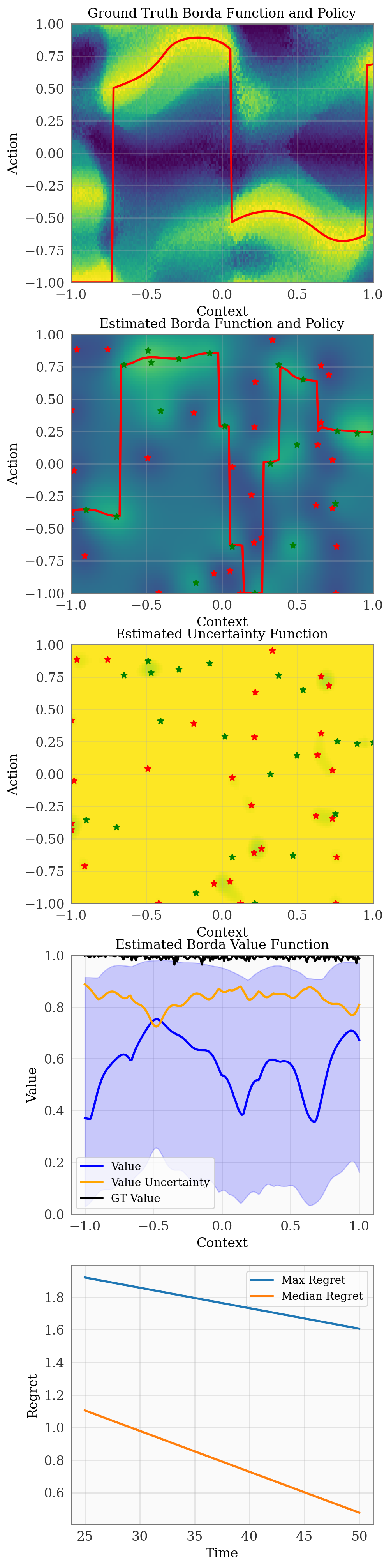}
    \includegraphics[width=0.3\textwidth]{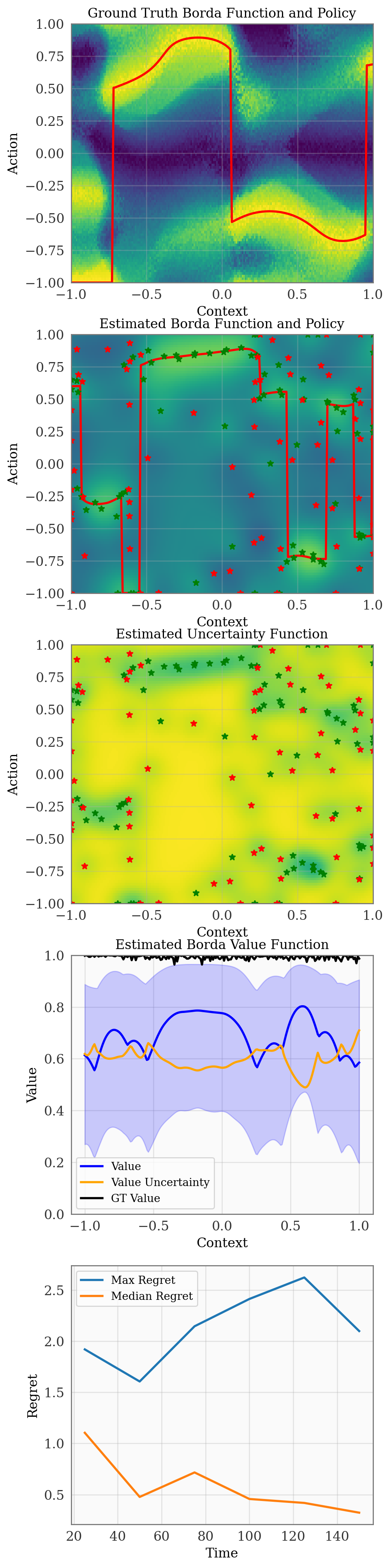}
    \includegraphics[width=0.3\textwidth]{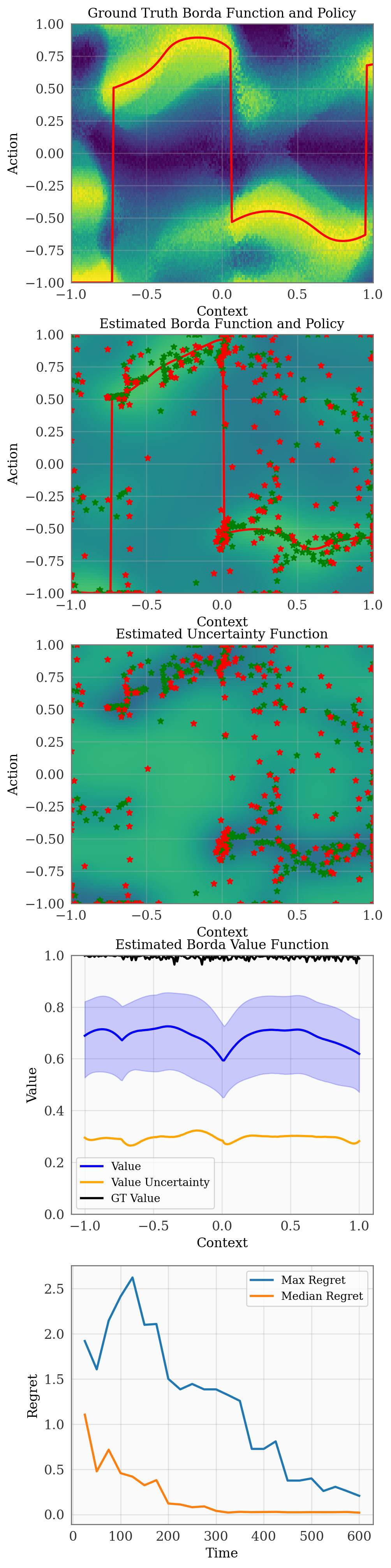}
    \vspace{-3mm}
    \caption{Progress of \algnm~across 50, 150, and 600 datapoints. From the top the charts show the ground truth function, the mean of the posterior estimate of $f_r$, the uncertainty function, the estimate of the value function as well as the acquisition function given in \eqref{eq:context_selection}, and the regret.}
    \label{fig:progression}
\end{figure*}

In order to assess the validity of our theory we have begun an experimental campaign starting with synthetic experiments that allow us to come as close as possible to the theoretical setting and empirically confirm our results.
To do so, we implemented the regression using the \textrm{BernoulliGP} model provided by GPyTorch \citep{gardner2018gpytorch}. We use a Mat\'ern kernal with automatic relevance detection with hyperparameters fit via maximum a posteriori optimized by the Adam algorithm \citep{kingma2014adam}.
We tested on distributions of synthetic reward functions generated by sampling a random linear combination of Random Fourier Features \citep{rahimi_rff} derived from a squared exponential kernel.
For each sampled reward function $r$, we used the Bradley-Terry model where $p(a  \preferred a'\mid x) = \frac{1}{1 + \exp(r(x, a') - r(x, a))}$ to generate comparison data. 
For each trial we uniformly sampled $n_0=25$ datapoints and then selected data to observe until $T=500$ total datapoints had been collected according to one of three methods:

\begin{figure}[t]
    \centering
    \includegraphics[width=0.45\textwidth]{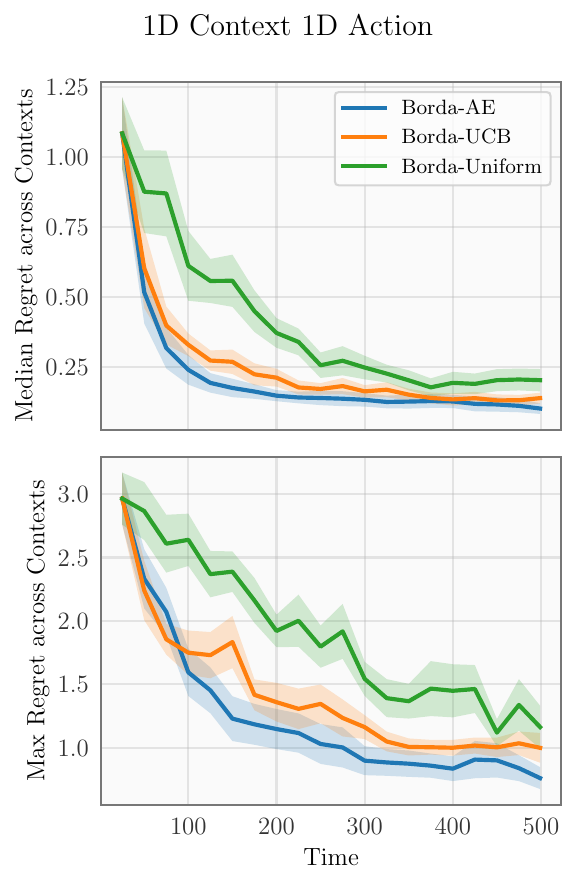}
    \vspace{-4mm}
    \caption{Performance of all methods across 10 random functions $r$ with 1D Context and 1D action. The top plot shows the median regret across contexts and the bottom shows the maximum. Error regions are standard errors.}
    \label{fig:comparison}
    \vspace{-5mm}
\end{figure}
\vspace{-3mm}
\begin{itemize}
    \item \textbf{\algnm}: our method, as described in Section~\ref{sec:methodandanalysis}.
    \item \textbf{\uniform}: uniform sampling of context and actions.
    \item \textbf{\uucb}: uniform sampling of contexts with UCB actions as in \algnm.
\end{itemize}
\vspace{-3mm}
This last method reduces to the method presented in \citet{xu2020zeroth} when na\"ively generalized to the contextual setting. All methods have the same test-time behavior of executing the action found by optimizing the pessimistic Borda function estimate for the test context.
By optimizing the ground-truth reward function we were able to approximate the optimal policy and therefore estimate the regret of our policy against it.
We give an example of the progression of our method for 1D context and 1D actions in Figure~\ref{fig:progression} as well as a comparison against \uniform~and \uucb~in Figure~\ref{fig:comparison}. One can see that \algnm~performs best both on median regret and on the maximum regret, which was the metric of interest in our theoretical analysis. 

It is clear the method is quickly able to concentrate samples in regions that could plausibly be the optimum and it is similarly clear that the peaks in the acquisition function over contexts are sensible given the mean and uncertainty estimates of $f_r$.

 \vspace{-1mm}
\section{Conclusion and Future Work}
\vspace{-1mm}
In this work we introduced the offline contextual dueling bandit setting and presented a first efficient algorithm for solving it in the kernelized setting. Though our method is able to in theory and in controlled practical settings achieve promising performance, it has a number of issues that prevent its wider application. First, we reflect on the fact that the Borda function is a blunt tool: in order to make sampling tractable we must sample $a'$ uniformly during exploration. Since we have the possibility of choosing both $a$ and $a'$ this seems somewhat wasteful of the $a'$ samples. In the context of RLHF with language models this is especially grim as uniformly chosen sequences are likely to be gibberish and the obvious preference for plausible sequences vs uniform ones will mean that the data selected via this strategy would be less likely to be useful. 

Many of these applications also benefit from the parallel nature of modern hardware and only make sense when presented with large batches of data. The sequential nature of \algnm~makes it unsuitable for these applications as well. 

With all this in mind, however, we still believe that there is tremendous potential in finding methods for solving these problems, especially with the worst-case style guarantees such as the ones provided here. We hope to continue the progress as we work towards solving these issues. 
\bibliography{main}
\bibliographystyle{icml2023}

\newpage
\appendix
\onecolumn
\section{Proof of Theorem \ref{thm:regret}}
\label{a:regret_proof}
In this section we will prove our main Theorem, \ref{thm:regret}. The overall strategy of the proof is to use our Lipschitz assumption on the link function (more precisely, the relative Lipschitzness of the reward $r$ and the Borda function $f_r$) in order to go to the Borda function, which we can directly model from data. Then, we use our selection criteria as well as confidence bounds taken from \citet{chowdhury2017kernelized} and convergence rates taken from \citet{kandasamy2019multi} in order to complete the argument. We give these cited results as lemmas in what follows.

In order to attain a particular policy performance with probability $1 - \delta$, we must bound the error of the estimates given by our KRR process for a particular confidence level. 
In order to do so, we adapt the result from \citet{chowdhury2017kernelized}, Theorem 2.
\begin{lemma}
    \label{lem:confidence_bounds}
    Let $\bonus = 2||f_r||_{\kernel} + \sqrt{2(\Phi_{t-1}(\Contextspace) + 1 + \log(2 / \delta))}$.
    Then with probability $1 - \delta / 2$ we have for all time $t$ and any point $\xinset$,
    \[|\mu_{t-1}(x) - f_r(x)| \leq \bonus \sigma_{t-1}(x).\]
\end{lemma}
This lemma jointly bounds the modeling error over the Borda function for all time $t$ though it introduces a dependence on the RKHS norm of $f_r$. This dependence is inherited from prior work, but we empirically study the relationship between the RKHS norm of a particular reward function and that of the associated Borda function in Section \ref{s:rkhs_borda}. 

We also adapt a result from Lemma 8 of \citet{kandasamy2019multi} in order to understand the convergence of our uncertainty function $\sigma_t$.
\begin{lemma}
    \label{lem:convergence}
    Suppose we have $n$ queries $(q_t)_{t=1}^n$ taken from $\Contextspace \times \Actionspace$. Then the posterior $\sigma_t$ satisfies
    \[\sum_{q_t}\sigma^2_{t-1}(q_t)\leq \frac{2}{\log(1 + \eta^{-2})} \Phi_{n}(\Contextspace\times\Actionspace).\]
\end{lemma}
Lemma~\ref{lem:convergence} gives us a handle on how quickly we can expect the uncertainty function to shrink as additional datapoints are observed. 

Now that we have lemmas \ref{lem:confidence_bounds} and \ref{lem:convergence} in place, we can proceed to the proof of the main result.

\begin{proof}
        In this proof, we condition on the event in Lemma~\ref{lem:confidence_bounds} holding true. 
        Given that occurence, we can say the following for every $\xinset$.
        \begin{align}
                \max_{\ainset} r(x, a) - r(x, \bestpolicy(s)) & \overset{\text{Assumption \ref{ass:borda}}}{\leq} L_1
                \left(\max_{\ainset} f_r (x, a) - f_r(x,\bestpolicy (x))\right) \\
             & \overset{\text{Lemma~\ref{lem:confidence_bounds}}}{\leq}
                L_1\left(\max_{\ainset} f_r (x, a) -   \max_{t \in [T]}\; \lcbr(x, \bestpolicy(x))\right)  \\ 
             &\overset{ \text{Def. of } \bestpolicy}  {=} 
                 L_1\left(\max_{\ainset} f_r (x, a)  - \max_{\ainset} \max_{t \in [T]}\; \lcbr(x, a)\right) \\
             &= 
                 L_1\min_{t \in [T]} \left(\max_{\ainset} f_r (x, a)  - \max_{\ainset} \; \lcbr(x, a)\right) \\
             & \overset{\text{Lemma~\ref{lem:confidence_bounds}}} {\leq}
                 L_1\min_{t \in [T]} \left(\max_{\ainset} \ucbr(x, a) - \max_{\ainset} \; \lcbr(x, a)\right) \\
            & \overset{ \text{Def. of } x^t}{\leq} 
                L_1\min_{t\in [T]} \left(\max_{\ainset} \ucbr(x^t, a) - \max_{\ainset} \; \lcbr(x^t, a)\right) \\
            & \overset{\text{Def. of } a^t} {\leq}  
                L_1\min_{t \in [T]}\left( \ucbr(x^t, a^t) -  \; \lcbr(x^t, a^t)\right) \\
            & \leq 
                \frac{L_1}{T}\sum_{t=1}^T \left( \ucbr(x^t, a^t) -  \; \lcbr(x^t, a^t)\right)\\
            & = \frac{L_1}{T}\sum_{t=1}^T 2\bonus \sigma_t(x^t, a^t)\\
            & \overset{\bonus\text{ is increasing}}{\leq} \frac{2L_1\lastbonus}{T}\sqrt{\left(\sum_{t=1}^T\sigma_t(x^t, a^t)\right)^2}\\
            & \overset{\text{Cauchy-Schwarz}}{\leq}
                \frac{2L_1\lastbonus}{T}\sqrt{T\sum_{t=1}^T\sigma^2_t(x^t, a^t)}\\
            & \overset{\text{Lemma~\ref{lem:convergence}}}{\leq}
                \frac{2L_1\lastbonus}{\sqrt{T}}\sqrt{C_1 \Phi_T}\\
            & \overset{\text{def of }\lastbonus}{=} \frac{2L_1}{\sqrt{T}}(2B + \sqrt{2(\Phi_{t-1} + 1 + \log(2 / \delta))})\sqrt{C_1 \Phi_T}\\
            & = O \left(  \frac{L_1}{\sqrt{T}} \left(B + \Phi_T\sqrt{\log \frac{1}{\delta}} \right)\right). 
        \end{align}

\end{proof}

\section{RKHS norms of $r$ and $\borda$}
\label{s:rkhs_borda}
In order to understand the dependence of our estimation bound on the RKHS norm $||\borda||_{\kernel}$, we ran numerical experiments on sampled reward functions. For a variety of context and action dimensions, we sampled 1000 reward functions as in Section~\ref{s:experiments} and numerically approximated their RKHS norms. We also made a Monte-Carlo estimate of the Borda function $f_r$ for each of the reward functions sampled and numerically approximated its RKHS norm. 
To do this, we uniformly sample 1,000 points $x_i$ from the input space, compute the regularized kernel matrix $K$ for this set $x_i$, solve the KRR problem $K \alpha = f(x)$ for $\alpha$. Then we compute the quadratic form $\sqrt{\alpha^T K\alpha}$ as an estimate of the RKHS norm.

In Table~\ref{tab:borda_norm}, we present the results of comparing the RKHS norms of 1000 reward functions and their associated Borda functions sampled as in Section~\ref{s:experiments}. A `win' was counted when the Borda function had smaller RKHS norm and a `loss' otherwise. The win margin is the average difference in RKHS norms of the reward and Borda functions, with a positive value when the Borda function was of smaller norm. 
It is clear here that in 
general (though not always) the RKHS norm of the Borda function $\borda$ for a particular reward function $r$ is smaller than the RKHS norm of the reward function $r$ itself. This relationship seems to grow stronger as the input dimensionality of the reward function grows larger. 

\begin{table}[]
\centering
\begin{tabular}{@{}llll@{}}
\toprule
Context Dimension & Action Dimension & Win Rate & Win Margin\\ \midrule
0 & 1 & 0.16 & -6.3 \\ 
1 & 1 & 0.89 & 5.1 \\
1 & 3 & 1 & 21.4 \\
3 & 1 & 1 & 21.5 \\
3 & 3 & 1 & 38.7 \\
10 & 10 & 1 & 19.6 \\
\bottomrule
\end{tabular}
\caption{Comparison of RKHS norms of reward functions and associated Borda functions}
\label{tab:borda_norm}
\end{table}

\end{document}